\documentclass{article}
\usepackage{spconf,amsmath,graphicx}
\usepackage{url}
\usepackage{caption}
\usepackage{subcaption}
\usepackage{dblfloatfix} 
\usepackage{float}

\usepackage[absolute,showboxes]{textpos}

\TPMargin{5pt}

\newcommand{\copyrightstatement}{
    \begin{textblock}{0.84}(0.08,0.93)    
         \noindent
         \footnotesize
         \copyright  20XX IEEE. Personal use of this material is permitted. Permission from IEEE must be obtained for all other uses, in any current or future media, including reprinting/republishing this material for advertising or promotional purposes, creating new collective works, for resale or redistribution to servers or lists, or reuse of any copyrighted component of this work in other works.
    \end{textblock}
}

\title{Analysis of Visual Reasoning on One-Stage Object Detection}

\twoauthors
 {Tolga AKSOY}
	{ASELSAN Inc.\\
    tolga.aksoy@aselsan.com.tr}
 {Uğur HALICI}
	{Middle East Technical University\\
	halici@metu.edu.tr}
 
\begin{document}

\copyrightstatement

\maketitle

\begin{abstract}
Current state-of-the-art one-stage object detectors are limited by treating each image region separately without considering possible relations of the objects. This causes dependency solely on high-quality convolutional feature representations for detecting objects successfully. However, this may not be possible sometimes due to some challenging conditions. In this paper, the usage of reasoning features on one-stage object detection is analyzed. We attempted different architectures that reason the relations of the image regions by using self-attention. YOLOv3-Reasoner2 model spatially and semantically enhances features in the reasoning layer and fuses them with the original convolutional features to improve performance. The YOLOv3-Reasoner2 model achieves around 2.5\% absolute improvement with respect to baseline YOLOv3 \cite{yolov3} on COCO \cite{coco} in terms of mAP while still running in real-time. 
\end{abstract}
\section{Introduction}
\label{sec:intro}

Object detection aims to classify and localize objects of interest in a given image. It has attracted great attention of the community because of its close ties with other computer vision applications. Many traditional methods have been proposed to solve object detection problem before the major breakthrough in deep learning area. These methods \cite{viola1, viola2, hog, dpm1, dpm2} were built on handcrafted feature representations. Inevitable dependency on handcrafted features limited the performance of traditional approaches. The great impact of the AlexNet \cite{alexnet} has put a new complexion on the object detection approaches and then deep learning based methods have completely dominated literature. Deep learning based detectors can be divided into two categories: two-stage object detectors and one-stage object detectors. Two-stage detectors have low inference speeds due to the intermediate layer used to propose possible object regions. Region proposal layer extracts regions of objects in the first stage. In the second stage, these proposed regions are used for classification and bounding box regression. On the other hand, one-stage detectors could predict all the bounding boxes and class probabilities in a single pass with high inference speeds. This makes one-stage detectors more suitable for real-time applications.

Recent one-stage object detectors \cite{yolov1, ssd, yolov3, efficientdet, yolov4} achieve good performance on datasets such as MS COCO \cite{coco} and PASCAL VOC \cite{pascalvoc}. However, they lack of ability to consider possible relations between image regions. The current one-stage detectors treat each image region separately. They are unaware of distinct image regions due to small receptive fields when image size is considered. They depend solely on high-quality local convolutional features to detect objects successfully. However, this is not the way how human visual system works. Humans have an ability of reasoning to carry out visual tasks with the help of acquired knowledge. Many methods \cite{smn, relationnetwork, iterativevr, detr, vityolo} have been proposed to mimic human reasoning ability in object detection. On the other hand, these methods are mostly complicated and uses two-stage detection architectures. Thus, they are not applicable for real-time applications.

In this paper, we propose a new approach to incorporate visual reasoning into one-stage object detection. Different from ViT-YOLO \cite{vityolo}, we integrate multi-head attention based reasoning layer on top of the neck instead of the backbone. By this way, reasoning information about the relationships between different image regions can be extracted by using more meaningful, fine-grained and enhanced feature maps. 

This paper's contributions can be summarized as follows:

\begin{itemize}
\item We present that one-stage object detection can be improved by visual reasoning. A new architecture is proposed that can extract semantic relations between image regions to predict bounding boxes and class probabilities. 
\item We analyze the effect of using only reasoning features on object detection performance. We demonstrate that fusing backbone output only-convolutional and reasoning features achieves the best performance improvement over the baseline model while still running in real-time. 
\item We analyze the effect of utilizing reasoning on average precision improvement for each object category.
\end{itemize}

\begin{figure*}[ht]
	\centering
	\includegraphics[width=0.72\textwidth]{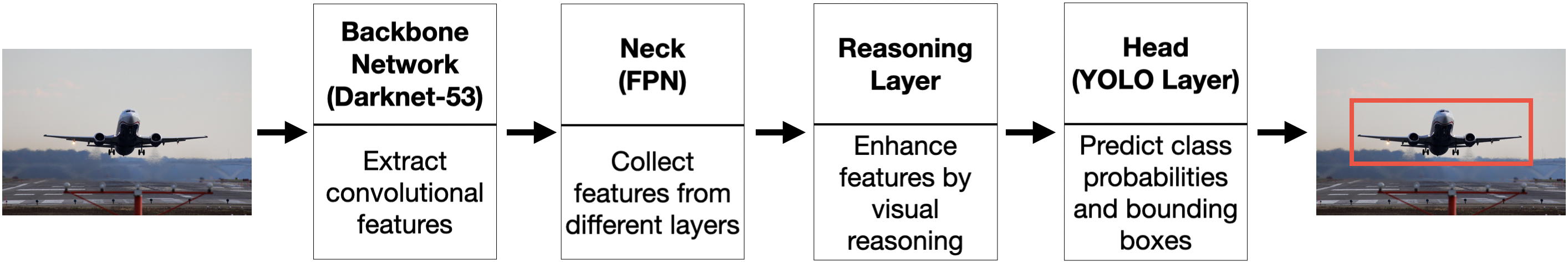}
	\caption{Overview of our approach.}
	\label{fig:overall_arch}
\end{figure*}

\section{Approach}
The general structure of the proposed method is shown in Figure \ref{fig:overall_arch}. Firstly, convolutional features are extracted by the Darknet-53 \cite{yolov3} backbone. Like YOLOv3, proposed method produces bounding box predictions at 3 different scales. Feature maps from different layers of the backbone are collected and concatenated after the necessary upsampling operations by FPN \cite{fpn} like neck. Then, the semantic relationships between image regions are extracted in the reasoning layer. At the final stage, class probabilities and bounding boxes are predicted by the YOLO head. 

\subsection{Reasoning Layer}
\begin{figure}[ht]
	\centering
	\includegraphics[width=0.43\textwidth]{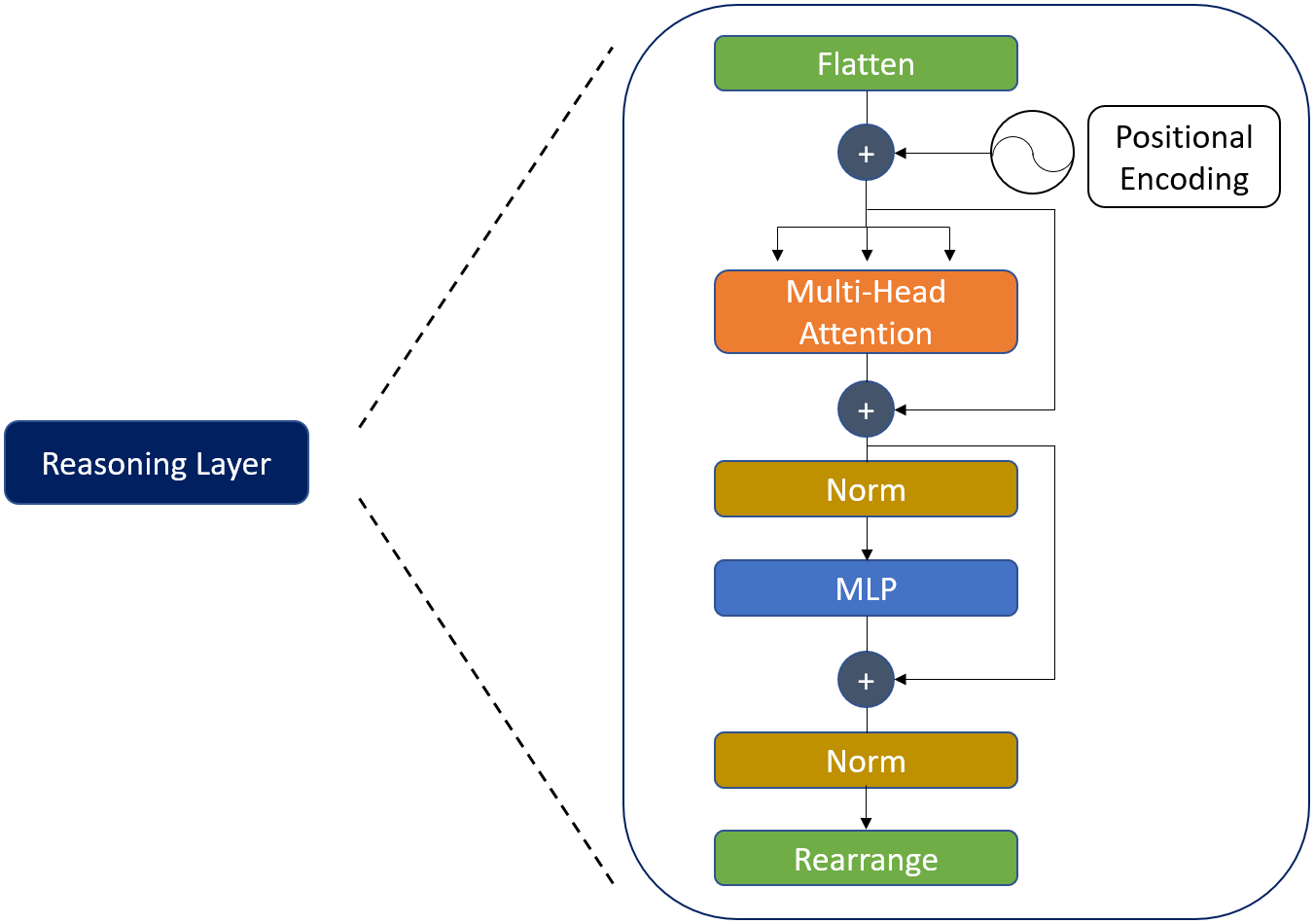}
	\caption{Reasoning layer}
	\label{fig:reasoning_layer}
\end{figure}

As a reasoning layer, transformer encoder-like \cite{attentionallneed} model is used. The architecture of the reasoning layer is shown in Figure \ref{fig:reasoning_layer}. All of the sub-layers of the reasoning layer are explained in detail in the following subsections. 

\subsubsection{Flatten}
The multi-head attention layer expects a sequence as an input. In Flatten, grid is converted to a sequence and fed to multi-head attention layer in this form.

\subsubsection{Positional Encoding}
By its nature, the multi-head attention layer is unaware of order in sequence. However, information about positions of the grid regions is valuable. To model order of image regions, fixed sinusoidal positional encoding \cite{pos_encod} is used:

\begin{equation}
{PE}_{\textit{(i, 2j)}} = \sin(\frac{\textit{i}}{10000^{{2j}/{d_{feature}}}})
\end{equation}

\begin{equation}
{PE}_{\textit{(i, 2j+1)}} = \cos(\frac{\textit{i}}{10000^{{2j}/{d_{feature}}}})
\end{equation}

where \textit{i} is the position of the grid region in the sequence, \textit{j} is the feature depth index, and \textit{d$_{feature}$} is the same with the feature depth. Generated values by sine and cosine functions are concatenated pairwise and added to convolutional feature embedding of the grid region.

\subsubsection{Multi-Head Attention}
Multi-head attention is the main layer where reasoning between grid cells, i.e. image regions, takes place. Reasoning between different regions of the input sequence is modeled by using self-attention which is based on three main concepts: query, key and value. In high level abstraction, query of a single grid cell in the sequence searches potential relationships and tries to associate this cell with other cells, i.e. image regions, in the sequence through keys. The comparison between query and key pairs gives us the attention weight for the value. Interaction between attention weights and values determines how much focus to place other parts of the sequence while representing the current cell.

In the self-attention, the query, key and value matrices are calculated by multiplying the input sequence \textit{X} with 3 different weight matrices: $W^Q$, $W^K$ and $W^V$:

\begin{equation}
Q = XW^Q
\end{equation}

\begin{equation}
K = XW^K
\end{equation}

\begin{equation}
V = XW^V
\end{equation}

To compare query and key matrices, the scaled dot-product attention is used \cite{attentionallneed}:

\begin{equation}
Attention(Q, K, V) = softmax(\frac{QK^T}{\sqrt{d_k}})V
\end{equation}

Each grid cell, i.e. image region, is encoded by taking a summation of attention weighted value matrix columns. The attention weights tell where to look in the value matrix. In other words, they tell which parts of the image are valuable, informative and relevant while encoding the current grid.

The self-attention mechanism was further improved by a multi-headed manner \cite{attentionallneed}. In the multi-head attention, self-attention computation is performed for a defined number of heads in parallel. The major superiority of multi-head over a single head is that it enables the model to work on different relationship subspaces. Each head has a different query, key and value matrices since each of these sets are obtained by using separate and randomly initialized weight matrices. The attention in head \textit{i} is calculated as:

\begin{equation}
h_i = Attention(QW_i^Q, KW_i^K, VW_i^V)
\end{equation}

Then, attentions are concatenated and transformed using a weight matrix $W^O$:

\begin{equation}
MultiHead(Q, K, V) = Concat(h_1, ..., h_n)W^O
\end{equation}

\subsubsection{Skip Connections}
There are two skip connections in the reasoning layer. Backpropagation is improved as stated in ResNet \cite{resnet} paper and original information is propagated to the following layers by residual skip connections.

\subsubsection{Normalization}
Normalization is applied in two places in the reasoning layer. Besides residual skip connection, normalization is another key factor to improve backpropagation. To deal with internal covariate shift \cite{batch_norm}, layer normalization \cite{layer_norm} is utilized.

\subsubsection{MLP}
Output of the multi-head attention is fed to multilayer perceptron (MLP) after normalization. MLP layer is composed of two linear layers and ReLU non-linearity in between:

\begin{equation}
MLP(x) = max(0, xW_1 + b_1)W_2 + b_2
\end{equation}

\subsubsection{Rearrange}
Rearrange is the last sublayer of the reasoning layer where sequence is converted back to the grid format which detection head expects.

\begin{figure*}[ht!]
	\centering
	\begin{subfigure}[b]{0.35\textwidth}
	    \centering
    	\includegraphics[width=\textwidth]{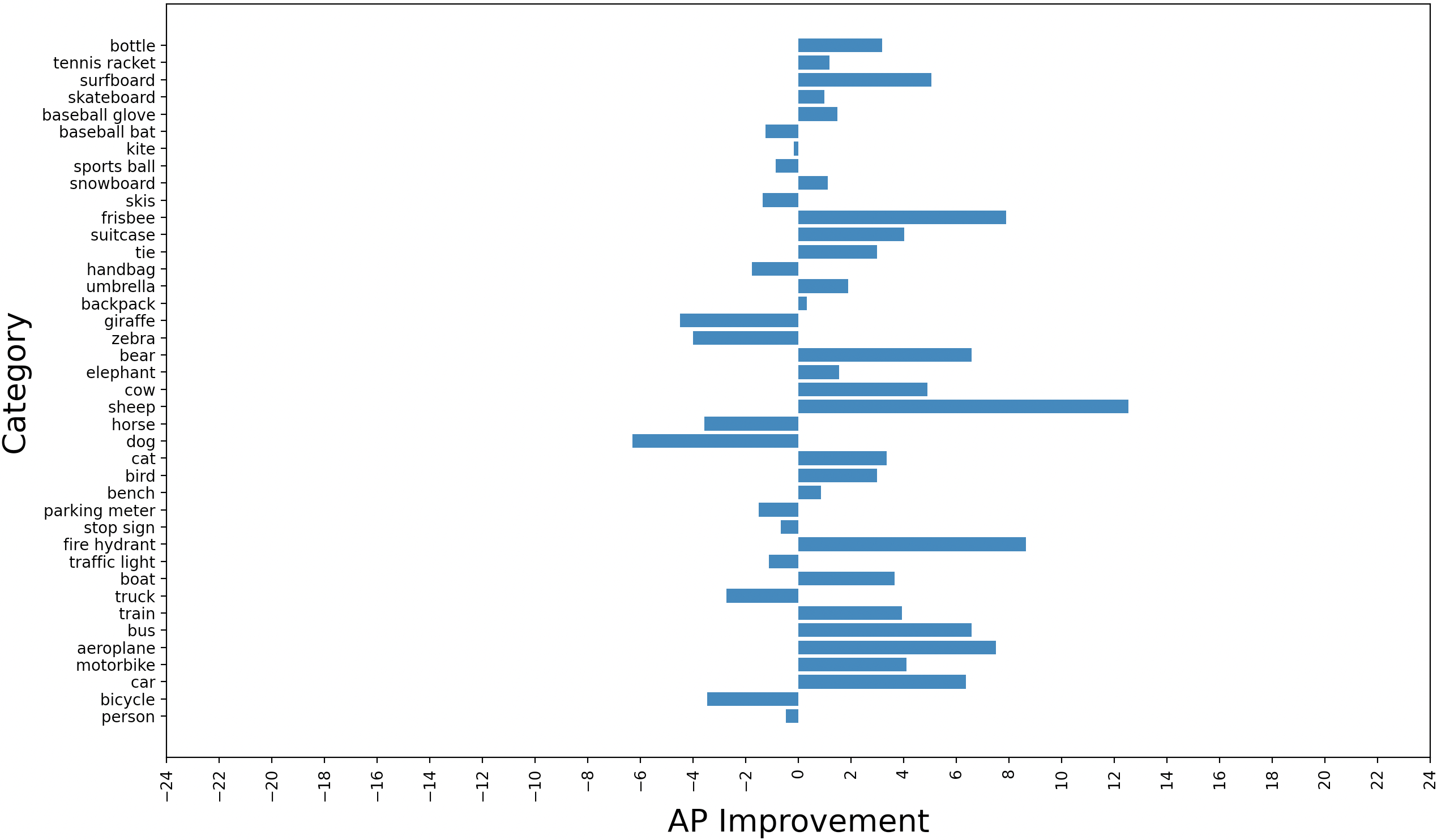}
    	\caption{YOLOv3-Reasoner1 first 40 categories}
    	\label{fig:arch}
	\end{subfigure}
	\begin{subfigure}[b]{0.35\textwidth}
	    \centering
    	\includegraphics[width=\textwidth]{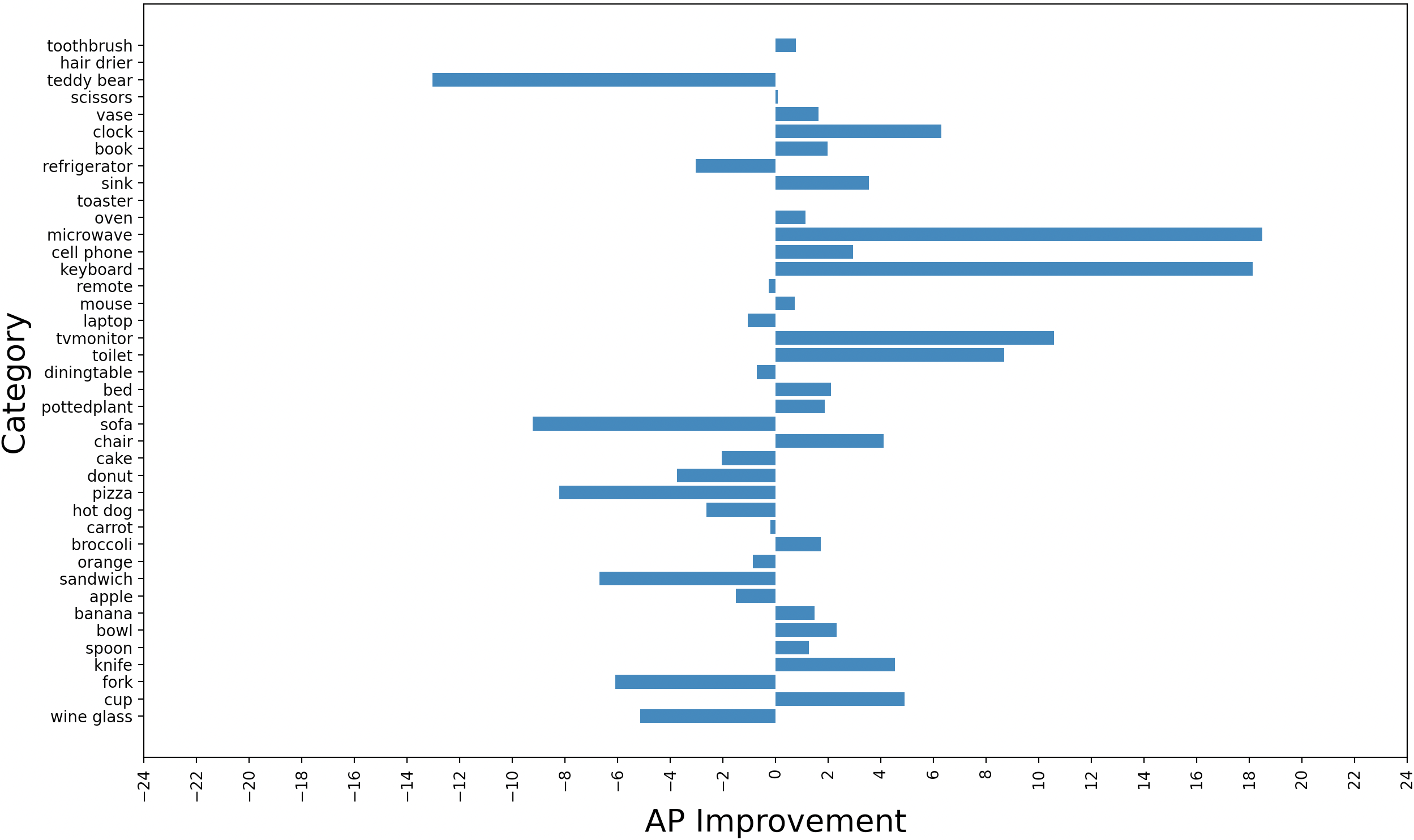}
    	\caption{YOLOv3-Reasoner1 last 40 categories}
    	\label{fig:arch}
	\end{subfigure}
	\begin{subfigure}[b]{0.35\textwidth}
	    \centering
    	\includegraphics[width=\textwidth]{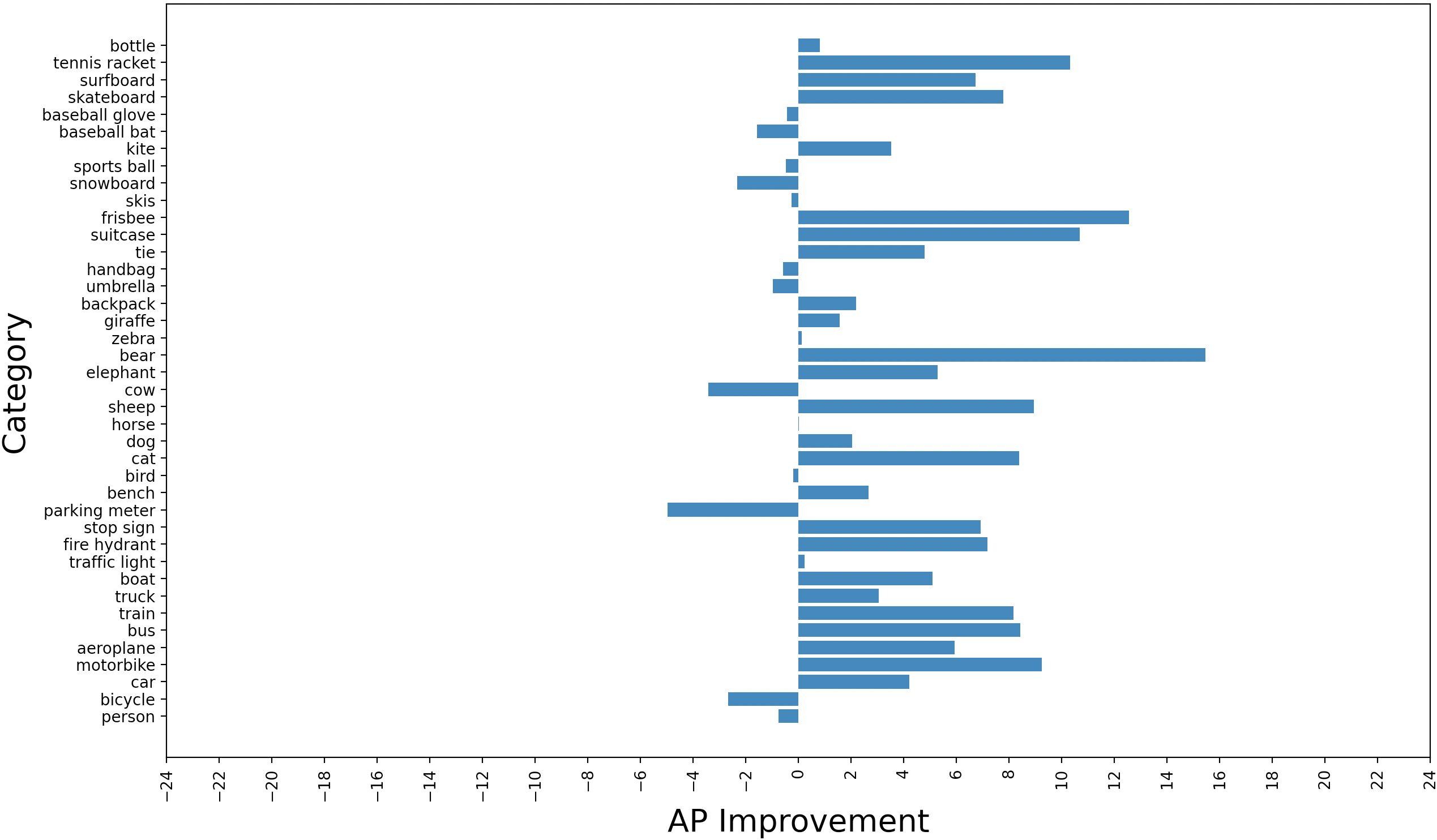}
    	\caption{YOLOv3-Reasoner2 first 40 categories}
    	\label{fig:arch}
	\end{subfigure}
	\begin{subfigure}[b]{0.35\textwidth}
	    \centering
    	\includegraphics[width=\textwidth]{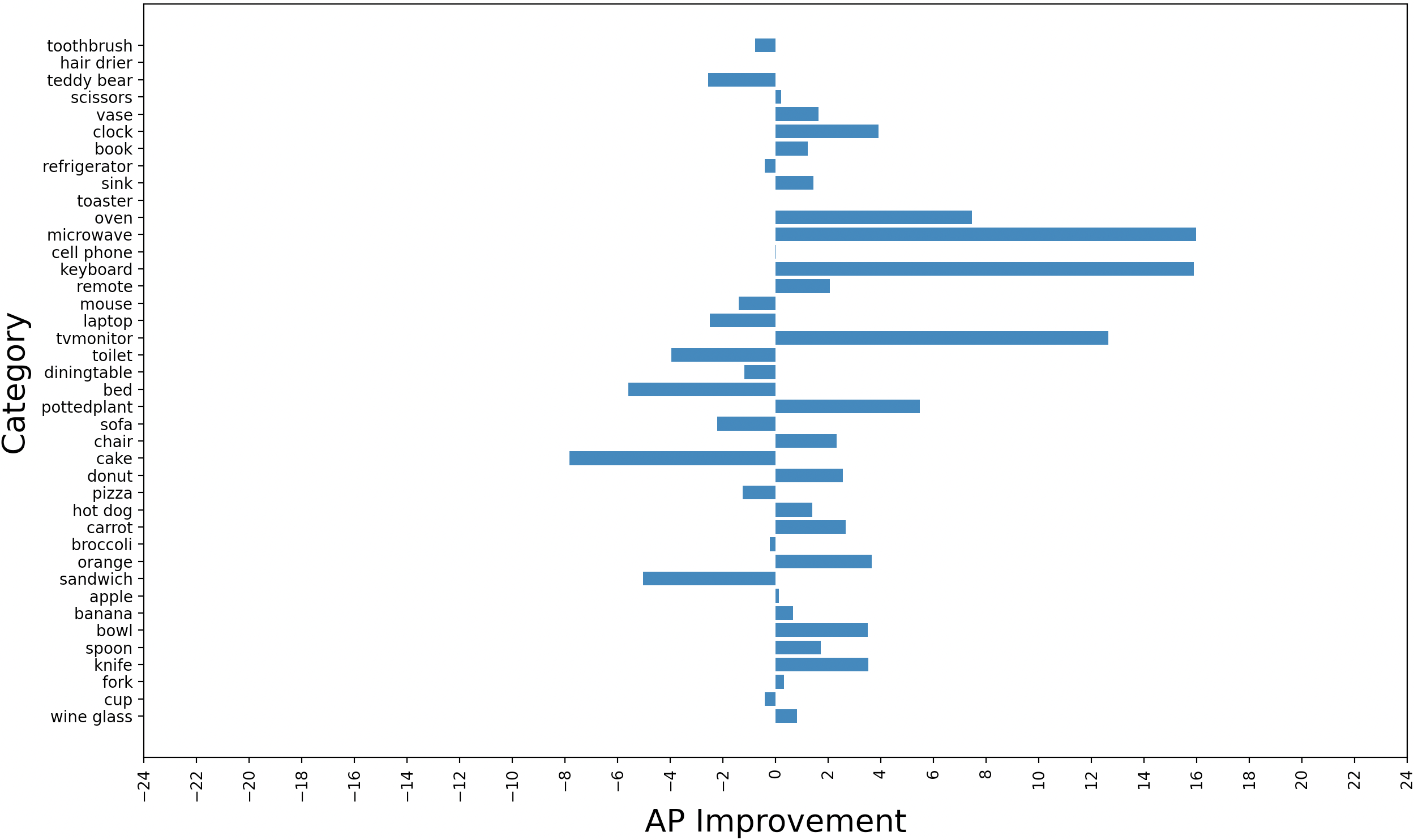}
    	\caption{YOLOv3-Reasoner2 last 40 categories}
    	\label{fig:arch}
	\end{subfigure}
    \caption{Difference between reasoner networks and the baseline YOLOv3 in terms of AP for each category of the COCO dataset }
    \label{fig:class_base_results}
\end{figure*}

\subsection{Reasoner Configurations}
Two different network configurations are trained and tested on COCO \cite{coco} dataset. In the first configuration, only reasoning features are fed to the detection head. In the second configuration, backbone output convolutional features are concatenated with reasoning features. Then, this new feature set is fused in 1x1 convolutional layer and fed to the detection head. Details related with network configurations are given in the following subsections. 

\subsubsection{YOLOv3-Reasoner1}
\begin{figure}[H]
	\centering
	\includegraphics[width=0.32\textwidth]{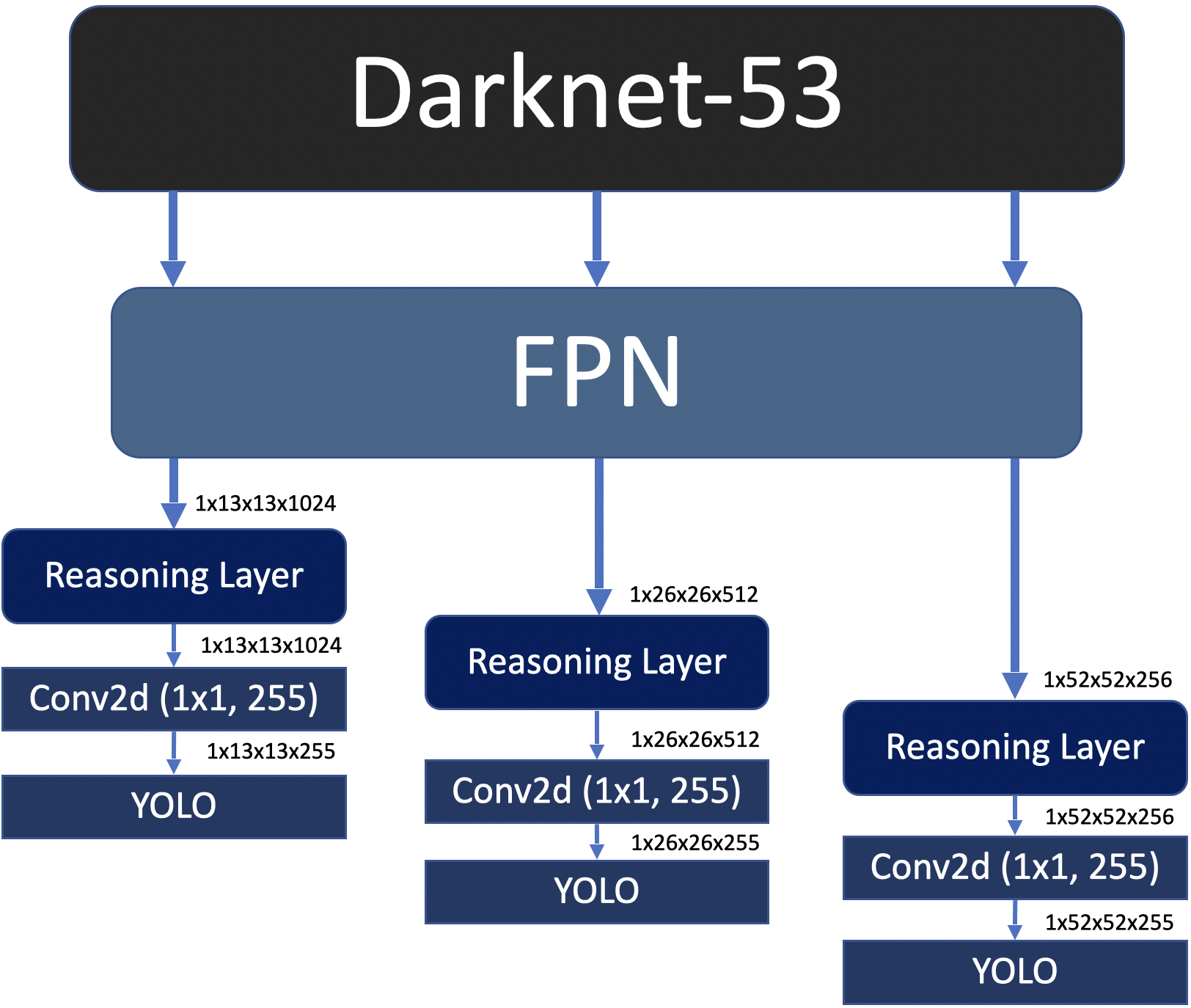}
	\caption{YOLOv3-Reasoner1}
	\label{fig:yolov3_reasoner1}
\end{figure}

FPN output is directly fed to the reasoning layer in this configuration. Number of heads are chosen 16, 8 and 4 for each scale respectively so that embedding size for each head has become 64. The reasoning layer output is fed to 1x1 convolutional layer. The whole architecture of YOLOv3-Reasoner1 is shown in Figure \ref{fig:yolov3_reasoner1}.

\subsubsection{YOLOv3-Reasoner2}
\begin{figure}[H]
	\centering
	\includegraphics[width=0.32\textwidth]{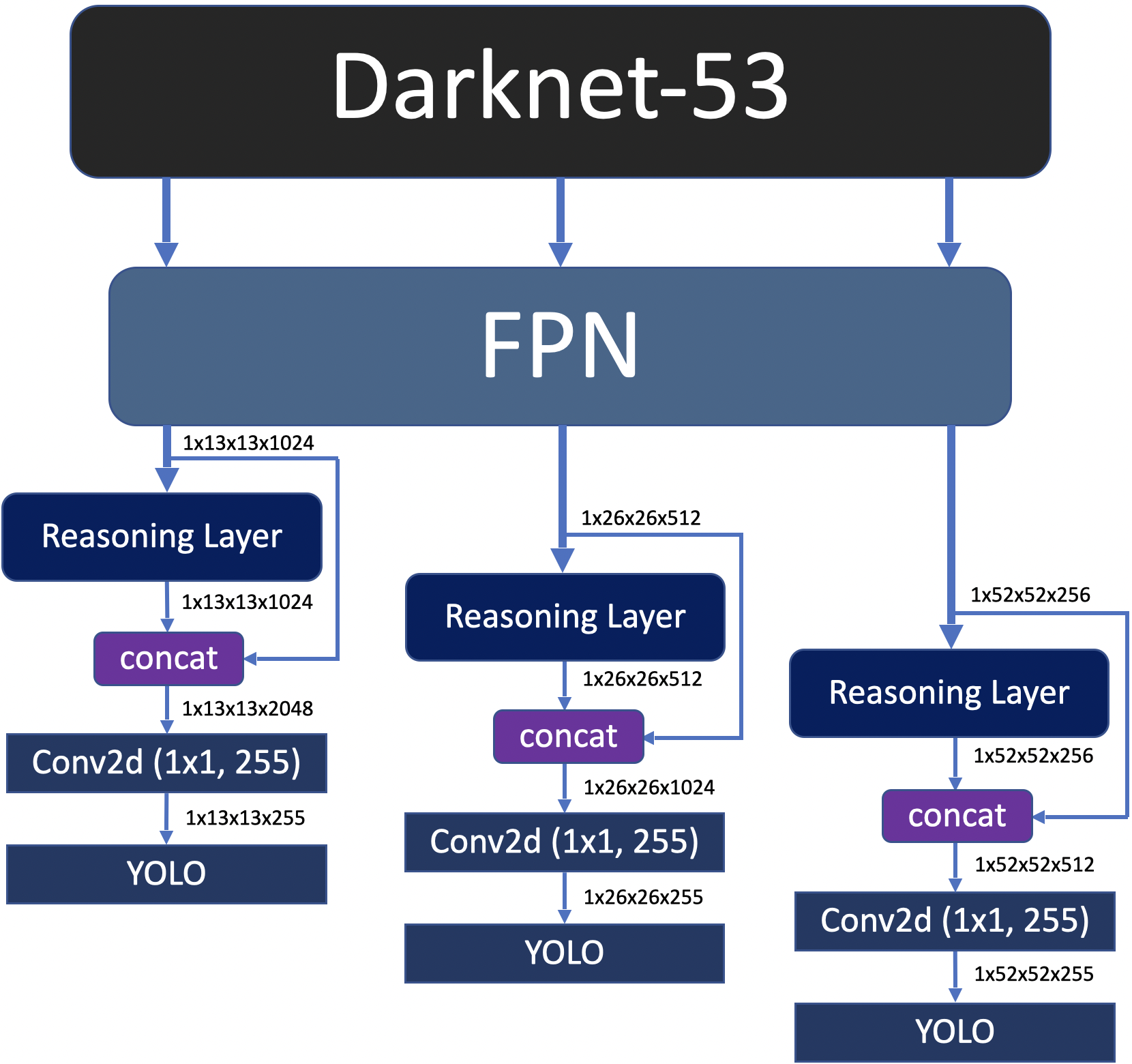}
	\caption{YOLOv3-Reasoner2}
	\label{fig:yolov3_reasoner2}
\end{figure}

In this configuration, output of the reasoning layer is concatenated with FPN output through a shortcut connection. Then, the output of the concatenation layer is fed to the 1x1 convolutional layer in order to fuse the information which is composed of reasoning and original only-convolutional features. There is a possibility that some parts of the convolutional features have been weakened in the reasoning layer. Our concatenation strategy ensures the reusability of the original convolutional features. The architecture of YOLOv3-Reasoner2 is shown in Figure \ref{fig:yolov3_reasoner2}. 

\section{Experiments}

In this section, we evaluate our reasoner networks. Firstly, dataset and evaluation metrics are introduced. Then, implementation details are given. Finally, quantitative and qualitative evaluation results are given.

\subsection{Dataset and Evaluation}
Experiments are performed on the MS COCO \cite{coco} dataset. The 2017 configuration of the dataset consists of 118K training and 5K validation images from 80 different object categories. As an evaluation metric, mean average precision (mAP) at IoU (Intersection over Union) = .5 is used as it is in the original YOLOv3 paper \cite{yolov3}.

\subsection{Implementation Details}
Network configurations are trained for 100 epochs from scratch on the COCO dataset. Initial learning rate was set to 0.001. Adam optimizer is used with parameters set to $\beta_1$ = 0.9, $\beta_2$ = 0.999 and $\epsilon$ = 1e-08. 

\subsection{Quantitative Evaluation}

In Table \ref{table:scores}, comparison results of our reasoner networks and the baseline YOLOv3. Frame rate comparison results are obtained by using Quadro RTX 8000 GPU. YOLOv3-Reasoner2 configuration where both the backbone output only-convolutional and reasoning features are used together achieves the best result while still running in real-time.

\begin{table}[h]
    \centering
    \caption{\textsc{Evaluation Metric Scores}}
    \label{table:scores}
    \begin{tabular}{c c c c}
        \hline
        Model            & \#Param. & FPS            &  mAP    \\
        \hline
        YOLOv3           & 61M      & \textbf{51}    & 26.18 \\
        \hline
        YOLOv3-Reasoner1 & 65M      & 46             & 27.45 \\
        \hline
        YOLOv3-Reasoner2 & 66M      & 45             & \textbf{28.68} \\
        \hline
    \end{tabular}
\end{table}

\subsection{Qualitative Evaluation}
Differences in average precision for each category of the COCO dataset between the reasoner configurations and the baseline YOLOv3 are examined. Results are shown in Figure \ref{fig:class_base_results}. The performance is improved on much of the categories. Improvements in YOLOv3-Reasoner2 are much better than the YOLOv3-Reasoner1. It seems that usage of the fpn output convolutional and reasoning features together lowers the performance degrade in some of the categories.

\section{Conclusion}
The results and analysis indicate that visual reasoning is promising to advance one-stage object detection. Although direct usage of the reasoning features in detection head achieves better performance than the baseline, fusing only-convolutional and reasoning features gives the best result by ensuring the reusability of the original backbone output convolutional features. For future work, the idea behind the YOLOv3-Reasoner2 model could be applied to a more recent one-stage object detection architecture to indicate new state-of-the-art.

\bibliographystyle{IEEEbib}
\bibliography{refs}

\end{document}